\documentclass{article}
\usepackage{ijcai25}

\usepackage{times}
\usepackage{soul}
\usepackage{url}
\usepackage[hidelinks]{hyperref}
\usepackage[utf8]{inputenc}
\usepackage[small]{caption}
\usepackage{graphicx}
\usepackage{amsmath}
\usepackage{amsthm}
\usepackage{booktabs}
\usepackage{algorithm}
\usepackage{algorithmic}
\usepackage[switch]{lineno}

\title{Claim Verification in the Age of Large Language Models: A Survey}

\author{
Alphaeus Dmonte$^1$ 
\and
Roland Oruche$^2$\and
Marcos Zampieri$^1$\and
Prasad Calyam$^2$\and
Isabelle Augenstein$^3$\\
\affiliations
$^1$George Mason University, USA\\
$^2$University of Missouri-Columbia, USA\\
$^3$University of Copenhagen, Denmark\\
\emails
admonte@gmu.edu
}
\begin{document}

\maketitle

\begin{abstract}
    The large and ever-increasing amount of data available on the Internet coupled with the laborious task of manual claim and fact verification has sparked the interest in the development of automated claim verification systems.\footnote{We use the terms \emph{claim verification} and \emph{fact verification} interchangeably given the overlap between the two concepts.} Several deep learning and transformer-based models have been proposed for this task over the years. With the introduction of Large Language Models (LLMs) and their superior performance in several NLP tasks, we have seen a surge of LLM-based approaches to claim verification along with the use of novel methods such as Retrieval Augmented Generation (RAG). In this survey, we present a comprehensive account of recent claim verification frameworks using LLMs. We describe the different components of the claim verification pipeline used in these frameworks in detail including common approaches to retrieval, prompting, and fine-tuning. Finally, we describe publicly available English datasets for this task.
\end{abstract}

\section{Introduction}
False information is widely present on social media and on the Web motivating the development of automated fact verification systems \cite{guo2022survey}. The introduction of LLMs has provided malicious actors with sophisticated ways of creating and disseminating false information. Recent election cycles has seen a large number of claims spread across social media and news platforms alike \cite{dmonte2024classifying}. Similarly, during the COVID-19 pandemic, many factually inaccurate claims were spread \cite{zhou2023synthetic}. 

\begin{figure}[!t]
    \centering
    \includegraphics[width=.75\linewidth]{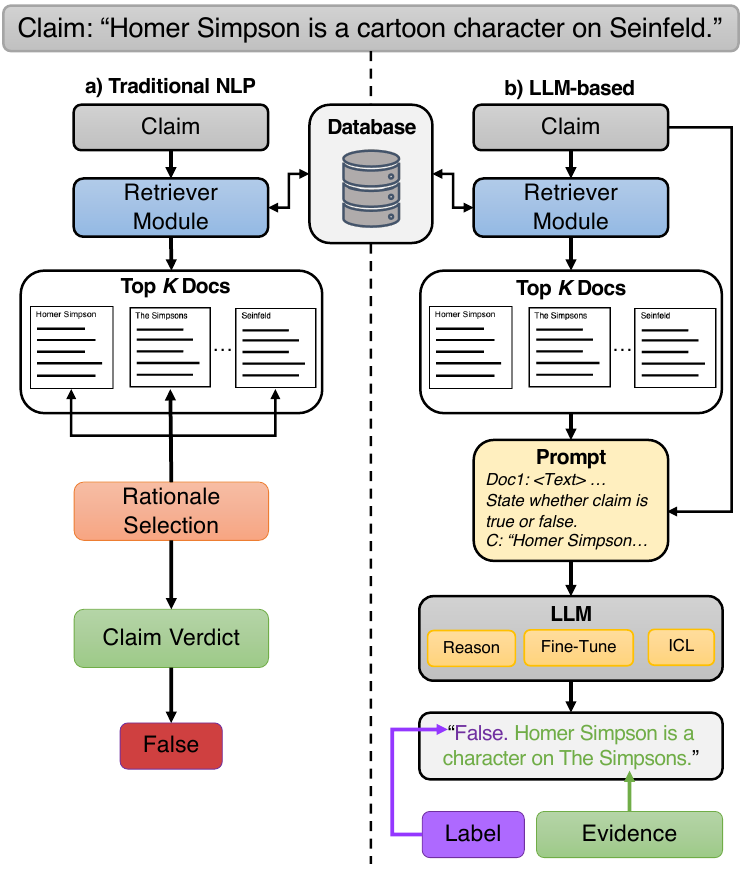}
    \caption{Comparison of claim verification systems between NLP-based (traditional) and LLM-based for claim veracity.}
    \label{fig:comparison}
\end{figure}

Fact-checking potentially false claims is an essential moderation task to reduce the spread of misinformation. Organizations such as FactCheck, PolitiFact, NewsGuard, and Full Fact perform manual fact-checking to verify claims in different domains. However, this is a laborious task requiring domain expertise \cite{adair2017progress,hanselowski2020machine}, which has become increasingly infeasible due to the sheer volume of misinformation generated by humans and AI models. Automated fact-checking has become an increasingly popular approach to verify the veracity of claims in a given text.\footnote{We acknowledge that claim verification models can also be applied to other modalities of data (e.g., images). In this survey, we address models applied to text only.} There are several steps involved in the fact-verification pipelines; the three main components are claim detection, evidence retrieval, and veracity prediction. 
Models used in these steps have followed the general methodological developments of the field and we have thus observed an increase in the use of LLMs for claim verification \cite{zhang2023towards,wang2023explainable,quelle2024perils}. Figure~\ref{fig:comparison} shows an example claim veracity scenario and compares the architectural differences between traditional NLP-based vs LLM-based claim verification systems. The latter are less prone to error propagation, and provide additional justifications when verifying claims.

Despite this, LLMs are pre-trained on very large text collections texts and are prone to hallucinations, often generating texts containing incorrect information. Further, such models can be used to generate misinformation at scale \cite{chen2023combating,zhou2023synthetic,dmonte2024classifying} and can therefore be exploited by malicious actors to spread factually incorrect information at an unprecedented rate \cite{pan2023risk}. Furthermore, using these pre-trained models for fact-verification may generate incorrect veracity labels, as the models may also rely on obsolete information. Approaches like RAG \cite{gao2023retrieval} are used for the model to retrieve the most recent information during fact verification.

A few general automated claim verification surveys have been published  \cite{zeng2021automated,bekoulis2021review,guo2022survey} as well as a couple of surveys on particular aspects of the task such as explainability \cite{vallayil2023explainability} and applications to specialized domains such as scientific texts \cite{vladika2023scientific}. However, all past related surveys lack consideration for LLM-based approaches or focus on specific sub-tasks of the pipeline \cite{panchendrarajan2024claim}. In this paper, we fill this important gap by surveying LLM-based frameworks proposed in recent years. To the best of our knowledge, this is the first survey to explore claim verification with LLMs. %We expect it to be a valuable resource to researchers in the field opening exciting new avenues for future research. %TODO includes some of the main findings here. 

% The remainder of this paper is structured as follows. Section \ref{sec:task-description} introduces automated claim verification and describes the multiple sub-tasks of a typical claim verification pipeline. We describe the datasets created for claim verification in Section \ref{sec:datasets} and the shared tasks organized on this topic in Section \ref{sec:shared-tasks}. Section \ref{sec:llm-approaches} provides a detailed description of the LLM-based approaches including retrieval, prompting, and fine-tuning strategies. Finally, Section \ref{sec:conclusion} concludes this survey and presents open challenges and future directions for research.

\section{Search Criteria}\label{sec:search-criteria}

We search popular repositories\footnote{ACM: \textit{portal.acm.org}. IEEE Xplore: \textit{ieeexplore.ieee.org}. Scopus: \textit{scopus.com}. ACL: \textit{aclanthology.org}. Web of Science: \textit{isiknowledge.com}. Springer: \textit{link.springer.com}. ArXiv: \textit{arxiv.org}. CEUR: \textit{https://ceur-ws.org/}.} of scientific articles to collect the papers that serve as primary sources for this survey, using the terms \emph{LLMs}, \emph{claim verification}, \emph{fact verification} and related keywords. We focus primarily on the \textit{ACL Anthology}, \textit{ACM Digital Library}, \textit{IEEE Xplore}, and proceedings of related conferences such as \textit{AAAI} and \textit{IJCAI}. We further search on \textit{Scopus}, \textit{Springer Link}, \textit{Science@Direct}, and \textit{ArXiv}.

Based on our search terms, we identify over 100 papers, and filter them as follows: (i) since the scope of our review is text-based LLM claim verification systems, we omit papers related to multimodal approaches (e.g., text-to-images), and papers investigating other modalities (e.g., images, graph-based systems); (ii) since we focus on LLM-based systems for claim veracity, we omit papers on claim identification, claim detection, and/or detecting LLM-generated content. This results in a total of 49 papers related to LLM-based approaches for veracity labeling. The papers on topic LLM-based veracity labeling have been published primarily in the ACL Anthology and ACM Digital Library, but also in other repositories such as the IEEE Xplore. %We publicly release this curated list.

\section{Claim Verification Pipeline}\label{sec:task-description}
\begin{figure}[!t]
    \centering
    \includegraphics[width=.35\textwidth]{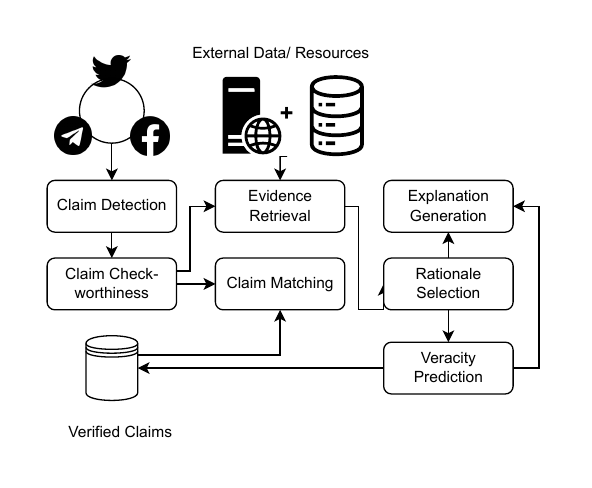}
    % \vspace{-0.3cm}
    \caption{A typical claim verification pipeline}
    \label{fig:pipeline}
\end{figure}

Figure \ref{fig:pipeline} shows a typical claim verification pipeline consisting of: claim detection, claim matching, claim check-worthiness, document/evidence retrieval, rationale/sentence selection, veracity label prediction, and explanation/justification generation. Many  systems only make use of some of these modules. %The following subsections describe each of them in detail. %Finally, as described in \cite{panchendrarajan2024claim}, performance evaluation of these sub-tasks is usually carried out using well-established automatic evaluation metrics such as Precision, Recall, and F-score. 

% \vspace{-0.6cm}

%\vspace{-0.5cm}

\paragraph{Claim Detection} Input texts may contain one or more statements, not all of which are claims. Given the input text, a claim detection module is designed to identify all the statements containing a claim. For example, the statement \emph{'I loved the movie Oppenheimer.'} is an opinion, whereas the statement \emph{'The COVID-19 pandemic started in Texas.'} contains a claim.

\paragraph{Check-Worthy Claim Identification} Not all  identified claims are check-worthy. Given an input claim, the task is to identify the claims that include real-world assertions and that may need to be verified \cite{hassan2015detecting,nakov2021clef}. Check-worthy claim identification is a subjective task and it relies on factors like popularity of the claim, public interest to determine the claim's veracity, etc. For example, the claim \emph{'The President met the State Governor to discuss the infrastructure deal.'} is less check-worthy than the claim \emph{'Drinking salt water cures COVID.'}

\paragraph{Claim Matching} The identified check-worthy claims can be matched to previously fact-checked claims. Given an input claim and a database of previously fact-checked claims, claim matching is used to determine if the input claim is previously fact-checked and exists in the database \cite{shaar2020known,nakov2021clef}. This can help predict the veracity label directly without additional steps.

\paragraph{Document/Evidence Retrieval} If the input claim does not exist in the database of fact-checked claims, it needs to be verified. In the evidence retrieval sub-task, all relevant documents related to the input claim are extracted, either from an external database or through an internet search~\cite{chen2017reading}. The threshold of the number of documents to be retrieved can be predetermined.

\paragraph{Rationale/Sentence Selection} Not all information in the retrieved documents is relevant to the claim. Hence for the rationale selection task, only the information or evidence most relevant to the claim is selected to predict the veracity label.

\paragraph{Veracity Label Prediction} Once a rationale is selected, it is provided to a classifier along with the claim and potentially additionally features, to predict a veracity label; often `SUPPORTED', `REFUTED', vs `NOT ENOUGH EVIDENCE'. For some datasets, the labels can be `TRUE' vs `FALSE'.

\paragraph{Explanation/Justification Generation} Recent works have focused on generating explanations for the veracity labels prediction. This specific task is focused on generating natural language justifications or explanations for the prediction, considering the claim and evidence.

\section{LLM Approaches}\label{sec:llm-approaches}
\begin{figure}[!t]
    \centering
    \includegraphics[width=1.05\linewidth]{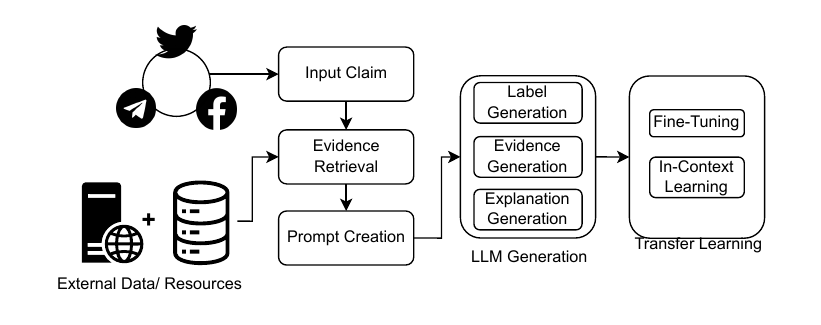}
    \vspace{-0.75cm}
    \caption{LLM-based claim verification pipeline. This involves creating a prompt from the retrieved evidence and the input claim as input to the LLM to generate a label, sentence evidence, and/or explanation of its response.}
    \label{fig:llm_pipeline}
\end{figure}

%In this section, we describe the recent advancements that enable LLMs to be robust in fact verification scenarios. 
Figure~\ref{fig:llm_pipeline} shows an example pipeline that encapsulates multiple component modules (i.e., Evidence Retrieval, Prompt Creation, Transfer Learning, and LLM Generation) for verifying claims. Different from traditional fact verification pipelines that select evidence sets for verifying claims, LLM-based claim verification conditions generated text based on the concatenated input claim and retrieved evidence. Retrieval-augmented LLMs have been shown to perform well on knowledge-intensive tasks such as text generation. %In the following, we detail the recent advancements of each component in relation to fact verification.

%\vspace{-0.45cm}

%\vspace{-0.4cm}

\subsection{Evidence Retrieval Strategies}
RAG models, which have been developed to address hallucination in LLMs for knowledge-intensive tasks, have shown success for fact verification~\cite{gao2023retrieval,guan2023language}. Early work such as~\cite{lewis2020retrieval} develop a framework that  retrieves evidence from an external database such as Wikipedia for conditionally generating veracity labels. Izacard et al.,~\shortcite{izacard2023atlas} demonstrate that RAGs perform well on the FEVER shared task~\cite{thorne2018fact}) in few-shot settings, showing 5\% improvement over large-scale LLMs such as Gopher~\cite{rae2021scaling} with significantly fewer parameters (i.e., 11B compared to 280B). Other works consider the optimization of either document ranking~\cite{glass2022re2g,chen2022corpusbrain} or input claims (i.e., queries)~\cite{hang2024trumorgpt} as a crucial step for improving evidence retrieval for veracity labeling. Hofst{\"a}tter et al.,~\shortcite{hofstatter2023fid} use an autoregressive re-ranker to get the most relevant passages from the retriever. These are then passed to the generation model to generate the veracity label.

Despite this, RAG models often fail when encountering long or complex input claims, resulting in incorrectly generated veracity labels or evidence sentences. Recent works have addressed this  by segmenting long claims into smaller sub-claims and performing multiple rounds of retrieval. Khattab et al.,~\shortcite{khattab2021baleen} presents a pipeline for multi-hop claim verification that uses an iterative retriever and neural methods for effective document retrieval and re-ranking. Shao et al.~\shortcite{shao2023enhancing} show that using a re-ranker to distill knowledge to a retriever helps close the semantic gaps between a query and document passage when verifying claims using iterative RAGs. Other works address the issue of complex claims using fine-grained retrieval techniques based on claim decomposition. Ches et al.~\shortcite{chen2023complex} generate sub-questions based on a claim, which a document retriever uses to retrieve relevant documents. A fine-grained retriever retrieves top-k text spans as evidence based on a k-word window and BM25. Zhang and Gao~\shortcite{zhang2023towards} decompose a claim into sub-claims and generate questions to verify the sub-claims. External knowledge sources are used to retrieve relevant information to verify the sub-claims and generate a final veracity label. Pan et al.~\shortcite{pan2023fact} follow a programming paradigm, where the claim is broken down into subtasks and the final label is the aggregation of the execution of each subtask. The fact-verification subtask uses external knowledge to retrieve relevant evidence for a claim.

Hang et al.~\shortcite{hang2024trumorgpt} retrieve evidence based on generated knowledge graphs of evidences. They generate a knowledge graph of the user query or input claim and compare it to the database of knowledge graphs to retrieve the most relevant information for claim verification. Hu et al.~\shortcite{hu2023give} propose a latent variable model that allows the retrieval of the most relevant evidence sentences from a document while removing the irrelevant sentences. This approach reduces noisy data during the verification process. Stochastic-RAG proposed by Zamani and Bendersky~\shortcite{zamani2024stochastic} uses a stochastic sampling without replacement process for evidence retrieval and selection. This approach overcomes the ranking and selection of the evidence hence optimizing the RAG model. Zhang et al.~\shortcite{zhang2023relevance} optimize the evidence retrieval process by using feedback from the claim verifier. The divergence between the evidence from a retrieved evidence set provided to the verifier and the gold standard evidence, acts as a feedback signal used to train the retriever. Xu et al.~\shortcite{xu2024search} propose Search-in-the-Chain, where an LLM generates a reasoning chain and based on the answer to each node in the chain, the retrieval can be used to correct an answer or provide additional knowledge, improving the generation accuracy of an LLM.

\subsection{Prompt Creation Strategies}
Text prompting has shown to be effective for improving the output of LLMs. In the context of claim verification, several works investigate both manual and automated prompting strategies to increase robustness. Zhang and Gao~\shortcite{zhang2023towards} develop a hierarchical prompting technique to verify multiple sub-claims using a step-by-step approach. Li et al.~\shortcite{li2023self} demonstrates a self-sufficient claim verification through prompting instructions on multiple language models. 

ProToCo~\cite{zeng2023prompt} demonstrates improved claim verification performance by leveraging a consistency mechanism to construct variants of the original claim-evidence pair prompt based on three logical relations (i.e., confirmation, negation, uncertainty). Chen et al.~\shortcite{chen2023unified} develop a unified retrieval framework that employs discrete, continuous, and hybrid prompt strategies for adjusting to various knowledge-intensive tasks such as claim verification. The FactualityPrompts~\cite{lee2022factuality} framework tests the output generations of LLMs given an input prompt and uses an external database such as Wikipedia to calculate factuality and quality measures compared to the ground truth. Other works aim to improve an LLM's reasoning abilities by appending the claims with evidence during prompting for claim verification and text generation~\cite{parvez2024evidence,dougrez2024assessing}. 

\subsection{Transfer Learning Strategies}

\paragraph{Fine-Tuning.} Although recent studies show the success of pre-trained LLMs on zero- or few-shot tasks, they often fail at verifying real-world claims given their limited internal knowledge. The success of fine-tuning has motivated recent work on claim verification. Chen et al.~\shortcite{chen2022gere} fine-tuned an LM on an external corpus for retrieving passage titles and evidence sentences using constrained beam search, finding improved performance on the FEVER dataset. Other work demonstrates that using GPT models to generate synthetic training data improves the performance of LLMs on fact checking~\cite{tang2024minicheck} and claim matching~\cite{choi2024automated}.

Pan et al.~\shortcite{pan2021zero} develop a pipeline for creating a fact verification dataset and fine-tuning a language model by leveraging passages from Wikipedia to generate QA pairs related to claim veracity, showing improved zero-shot performance. Zeng and Zubiaga~\shortcite{zeng2024maple} show that unlabelled pairwise data can increase the alignment between claim-evidence pairs, resulting in significant improvement on few-shot claim verification. Other recent works leverage reinforcement learning to fine-tune models for improving the veracity of claims and supporting evidence~\cite{zhang2024reinforcement,huang2024training}. A document-level and question-level retrieval policy is proposed by Zhang and Gao~\shortcite{zhang2024reinforcement}, where the top-k and top-1 documents for the document and question-level policy respectively are used as input to a scoring function for label prediction during training. This approach outperforms retrieval and prompting approaches. Chiang et al.~\shortcite{chiang2024team} fine-tune LLMs for multi-stage fact verification. They fine-tune a model to generate answers based on claim-evidence pairs and a set of questions, whereas another model is fine-tuned to verify the claim based on the claim-evidence and question-answer pairs. Zhu et al.~\shortcite{zhu2023explain} fine-tune a generation model to generate counterfactuals for out-of-domain classification. 

\paragraph{In-Context Learning.} The recent success of pre-trained LLMs in zero- and few-shot settings is largely attributed to its \textit{in-context learning} (ICL) abilities~\cite{kojima2022large,brown2020language}. For claim verification, popular ICL techniques include chain-of-thought (CoT) reasoning~\cite{wei2022chain}. Zhao et al.~\shortcite{zhao2024pacar} develop a multi-stage verification pipeline based on claim decomposition and self-reflection. An LLM-based verifier module is created using instruction prompting to generate a reasoning analysis among all sub-claims created by the decomposer module. The suggests that zero-shot prompting techniques result in better multi-hop performance on the HOVER and FEVEROUS datasets compared to few-shot prompting and fine-tuning. Kanaani~\shortcite{kanaani2024triple} enable LLMs to generate reasons over retrieved evidence in claim verification using few-shot ICL and the STaR CoT technique inspired by Zelikman et al.~\shortcite{zelikman2022star}. Similar work has leveraged CoT  for effectively verifying complex claims using reasoning steps~\cite{yao2023react,ni2024afacta}. Conversely, HiSS~\cite{zhang2023towards} demonstrates that prompting for few-shot learning and claim decomposition can substantially improve the performance of CoT models for complex news claim verification. Li et al.~\shortcite{li2023overprompt} leverage the ICL capability of LLMs to perform multiple tasks simultaneously. Their approach outperforms or achieves comparable task performance in a zero-shot setting on claim verification datasets.

\subsection{LLM Generation Strategies}
% Might be helpful to describe that not all claim verification pipelines use same approach for evidence selection and label prediction
\paragraph{Label and Evidence Generation.} While the majority of claim verification systems predict veracity  based on the concatenation of the input claim and evidence sentences~\cite{pradeep2021scientific}, recent work has proposed alternate strategies for determining veracity labels and selecting/generating evidence pieces. Cao et al.~\shortcite{cao2024can} develop SERIf, a claim verification pipeline that features an inference module to predict the veracity label of scientific news articles based on a two-step summarization (i.e., `Extractive - Abstractive`) and evidence retrieval technique. Each summary-evidence pair is fed into the LLM and produces a binary label, indicating whether the news article is reliable (supported) or unreliable (refuted). Wadden et al.~\shortcite{wadden2022multivers} leverage the Longformer model~\cite{beltagy2020longformer} that uses a shared encoding over the claim and document abstracts for rationale identification and claim label prediction. 

Li et al.~\shortcite{li2024minimal} propose to select the minimal evidence group within a set of retrieved candidate documents. This aims to minimize redundancy while selecting the most relevant evidence to prompt the language model. Chen et al.~\shortcite{chen2022gere} use BART to encode all candidate sentences from the most relevant retrieved documents. In this, BART serves as an evidence decoder to predict the $g$-th evidence sentence via generation conditioned on the top $k$ retrieved documents and the input claim. Lee et al.~\shortcite{lee2022factuality} develop a variant of nucleus sampling called \textit{factual-nucleus sampling}, in which the top-$p$ sampling pool is selected as a set of sub-words whose cumulative probability exceeds $p$, resulting in improved evidence and label generation without claim verification datasets such as FEVER. Kao and Yen~\shortcite{kao2024magic} propose a multi-stage approach, where the evidence sentences are retrieved from articles related to a claim, and arguments are generated by aggregating and reconstructing the evidence. The arguments are refined and passed to a LLM to generate a verification label. Other approaches leverage LLMs reasoning capabilities to generate veracity labels and factual evidence~\cite{cheng2024information,jafari2024robust,li2024chainofknowledge,fang2024counterfactual,pan2023qacheck}.

\paragraph{Explainable Generation.}
Recent studies investigate explainable approaches to improve LLM-based claim verification. Wang and Shu~\shortcite{wang2023explainable} present the FOLK framework, that leverages the explanation capabilities of LLMs when verifying claims and justifies the prediction through a summary of its decision process. Dammu et al.~\shortcite{dammu2024claimver} proposes a knowledge graph (KG)-based approach for text verification and evidence attribution, where the LLM is fine-tuned on evidence attribution based on the input text and retrieved triplets from KG, inducing explanations for claim predictions. While explanation techniques can help humans verify facts, LLMs can produce incorrect explanations due to hallucinations, making them unreliable in certain claim verification scenarios~\cite{si2023large}. Ma et al.~\shortcite{ma2023ex} prompt an LLM in a few-shot setting to generate a concise summary of the evidence documents and input claim, serving as an explanation for the verified claim. Other works leverage reasoning techniques such as chain-of-thought (CoT) to enable the LLM to be interpretable in its decision-making process when verifying claims~\cite{yao2023react,pan2023qacheck,zhao2024pacar,kanaani2024triple,ni2024afacta,quelle2024perils,fang2024counterfactual}. Pan et al.~\shortcite{pan2023qacheck} and Fang et al.~\shortcite{fang2024counterfactual} leverage an LLM's reasoning ability to generate explanations by using question-guided reasoning and minimizing the inherent model biases.
% Authors in~\cite{yao2023react} propose ReAct, an approach that appends the model reasoning to the current context, which is used for future reasoning. This approach is used interchangeably with the CoT approach for fact-verification. 
% An approach proposed by \cite{zhao2024pacar} uses a multi-stage verification pipeline. The pipeline relies on self-reflection and reinforcement learning. It uses a verifier module that uses the retrieved evidence and sub-claims to perform reasoning analysis to predict a veracity label and provide justification. 
% Authors in~\cite{kanaani2024triple} propose a framework called Triple-R consisting of a retriever, a ranker, and a reasoner module. The reasoner module is responsible for generating veracity labels along with a justification. The authors use in-context learning and the CoT approach to generate reasons for a claim. 
% The AFaCTA framework proposed by \citet{ni2024afacta} for factual claim annotation uses CoT to extract verifiable facts from the text. A reasoning justifying the extraction process is provided which aids in the final veracity label prediction.
% Authors in~\cite{} use Google search to retrieve contextual information that can aid the reasoning capability of an LLM to generate an explanation for a claims prediction.

\section{Evaluation and Benchmarking}

\subsection{Metrics}

The F1 score is the most commonly used metric to measure the performance of automatic claim verification systems, and to a lesser degree Precision, Recall, and Accuracy. Katranidis and Barany~\shortcite{katranidis2024faaf} used the error rate between the human and automated fact-verification system to measure verification accuracy. However, these metrics consider a single pipeline component to evaluate the system's overall performance. Hence, Thorne et al.~\shortcite{thorne2018fever} introduce the FEVER score, a metric that incorporates both verification accuracy and evidence retrieval accuracy to compute overall system performance. While these metrics are valuable for assessing the performance of the classification tasks, they are inadequate for evaluating the performance of the non-classification components of the pipeline. Hence, metrics like Recall@k are used to measure the performance of the retrieval task~\cite{pan2023fact,pradeep2021scientific}, while BLEU and METEOR are used to evaluate the quality of explanations or generated questions and answers. Schlichtkrull et al.~\shortcite{schlichtkrull2024averitec} propose the new evaluation metric AVeriTeC score that uses METEOR and accuracy for question-answer-based veracity prediction systems. Other metrics like Mean Absolute Error (MAE), Expected  Calibration Error (ECE), Area Under ROC Curve (AUC-ROC), and Pearson's Correlation are also used. Most of these metrics are inadequate for evaluating the factual accuracy of LLM-generated text, for which FactScore~\cite{min2023factscore}, SAFE~\cite{wei2024long}, and VERISCORE~\cite{song2024veriscore} have been proposed instead. Other metrics and frameworks evaluate factual errors in generated text~\cite{lee2022factuality,chern2023factool} and their alignment~\cite{zha2023alignscore} and entailment~\cite{lee2022factuality} considering factuality.

\subsection{Datasets}\label{sec:datasets}

The fundamental resource for training and evaluating claim verification systems is datasets containing annotated texts. As most research in this area deals with English data, we collect information about all publicly available English datasets used in the papers discussed in this survey\footnote{\url{https://github.com/LanguageTechnologyLab/Claim-Verification-Papers.git}}.
% and present them in Table \ref{tab:data}. 
 
General-domain datasets have been created from online data sources and websites including Wikipedia~\cite{thorne2018fever,jiang2020hover,diggelmann2020climate,eisenschlos2021fool,schuster2021get,kamoi2023wice}, due to the extensive amount of information available spanning various topics and domains, and online fact-checking websites like PolitiFact~\cite{wang2017liar,augenstein2019multifc,kao2024magic}. Factually incorrect claims are shared through social media channels like X, Reddit, etc. Saakyan et al.~\shortcite{saakyan2021covid} introduce the COVID-Fact dataset consisting of claims extracted from Reddit posts. Several datasets for scientific fact verification have also been introduced~\cite{wadden2020fact,wadden2022scifact,lu2023scitab}. Most fact-verification datasets rely on unstructured textual evidence. Hence, a few datasets with structured evidence sources have been introduced~\cite{aly2021feverous,lu2023scitab}. While many datasets focus only on veracity labels, some were developed to address explainability too~\cite{chen2022generating,yang2022coarse,ma2023ex,rani2023factify,schlichtkrull2024averitec}.

Most claim-verification datasets include claims extracted from available information sources. LLMs are widely used to generate content, even though they can generate factually incorrect information. Efforts to identify such non-factual text have been undertaken. However, claim-verification datasets consisting of human-written claims can be inadequate for this task due to the linguistic discrepancy between human- and LLM-generated text. As such, LLM-generated claim verification datasets have been introduced~\cite{li2023self,cao2024can}. While these are used to evaluate  automatic claim verification systems, there is a need to evaluate the factual accuracy of LLM-generated text, leading to the introduction of evaluation datasets~\cite{lee2022factuality,Wang2023FactcheckBenchFE,malaviya2024expertqa}.

\subsection{Shared Tasks}
Shared tasks are competitions where teams develop systems for tasks using a common benchmark dataset. Multiple shared tasks on claim and fact-checking  have been organized including the Fact Extraction and Verification (FEVER) challenge~\cite{thorne2018fact}, TabFact\footnote{https://competitions.codalab.org/competitions/21611}, CLEF 2020 CheckThat!~\cite{barron2020overview}, SCIVER~\cite{wadden2021overview}, SEM-TAB-FACT~\cite{wang2021semeval}, FACTIFY-5WQA\footnote{https://defactify.com/factify3.html}, and more recently AVeriTeC\footnote{https://fever.ai/task.html}. While there have been various techniques like question-answer generation as a precursor to label prediction as in AVeriTeC, all these shared tasks are centered on predicting veracity labels. Given an LLM's tendency to generate plausible yet factually inaccurate text, there is a need to organize shared tasks to evaluate the factual accuracy of LLM-generated content.

\section{Open Challenges}\label{sec:open-challenges}

\paragraph{Handling Irrelevant Context} Retrieved evidence may be irrelevant, which is a challenge for LLMs, as they may not be trained to ignore such evidence. The lack of robustness to  noise can lead an LLM to produce misinformation and incorrect verification. Recent research on open-domain question answering shows that external knowledge relevant to the task can improve model performance, however, irrelevant context can also lead to inaccurate predictions~\cite{petroni2020how,shi2023large,li2023large,yu2023chain}. For the fact-verification, recent work has proposed techniques to identify the most relevant context for improving veracity label prediction, thus improving the overall system performance~\cite{wang2023learning,yoran2024making,xia2024improving}. However, more work is required to verify the effectiveness across domains.
    
\paragraph{Handling Knowledge Conflicts} The reliance of fact-verification approaches on retrieved evidence can cause knowledge conflicts in LLMs, where retrieved external evidence may conflict with the internal parameters of the pre-trained LLM. This causes the LLM to ignore the retrieved evidence and produces hallucinations~\cite{xu2024knowledge}. Approaches for avoiding knowledge conflicts have been introduced for question answering ~\cite{li2023large,neeman2023disentqa,mallen2023not,longpre2021entity,chen2022rich,marjanovic2024knowledge}. Expanding this work to fact-verification, especially to the approaches that use LLMs, is vital for an effective verification process.

\paragraph{Multilinguality} Most automated claim verification approaches rely on English datasets. Furthermore, there are limited multilingual fact-verification datasets \cite{gupta2021x,kazemi2022matching,pikuliak2023multilingual,singh2023finding}. This hinders the development of approaches for multilingual fact-verification, which achieve the best performance when trained on language-specific datasets \cite{panchendrarajan2024claim}.

\section{Conclusion}\label{sec:conclusion}

We presented a survey on LLM approaches to claim verification. To the best of our knowledge, this is the first claim verification survey to focus exclusively on LLM approaches, thus filling an important gap in the literature. We have described the traditional claim verification pipeline including its component tasks and discussed various LLM-based approaches used in this task. Finally, we have also described publicly available English datasets providing important information to new and seasoned researchers on this topic. 

Advances in LLM development will likely continue to improve the quality of claim verification systems. We hope this survey motivates future research on this topic taking advantage of recently-proposed LLMs, RAG methods, and others. Claim verification is a vibrant research topic and we envisage multiple open research directions such as handling irrelevant context, knowledge conflicts and multilingualism.
% Claim verification is a vibrant research topic and we see multiple open research directions as described in section~\ref{sec:open-challenges}.

\bibliographystyle{named}
\bibliography{custom}

\end{document}